\documentclass[10pt,twocolumn,twoside] {IEEEtran}

\usepackage{amsmath,amssymb,amsfonts}
\usepackage{graphicx}
\usepackage{cite}
\usepackage{hyperref}
\usepackage{soul}
\usepackage{booktabs}
\usepackage[caption=false,font=footnotesize]{subfig}
\usepackage{orcidlink}

\usepackage{amsthm}
\newtheorem{theorem}{Theorem}
\newtheorem{corollary}{Corollary}
\theoremstyle{remark}
\newtheorem{remark}{Remark}

\title{Networked AI: Architecture, Theory, and Applications} 
\author{
Liang~Song\,\orcidlink{0000-0002-8143-9052},
Jiayue~Jin\,\orcidlink{0009-0007-5642-7846},
Peng~Sun\,\orcidlink{0000-0001-8356-1329},\IEEEmembership{Senior Member, IEEE},
Di~Li\,\orcidlink{0000-0002-7764-7609},
Yang~Liu\,\orcidlink{0000-0002-1312-0146},\IEEEmembership{Member, IEEE},
Athanassios~Skodras\,\orcidlink{0000-0002-3872-4325},\IEEEmembership{Life Senior Member, IEEE},
Dimitris~Ampeliotis\,\orcidlink{0000-0002-4138-6152},\IEEEmembership{Member, IEEE},
Yong~Yuan,
Amit~Dvir\,\orcidlink{0000-0002-3670-0784},
Jiangchuan~Liu\,\orcidlink{0000-0001-6592-1984},\IEEEmembership{Fellow, IEEE},
Victor~C.~M.~Leung\,\orcidlink{0000-0003-3529-2640}, \IEEEmembership{Life Fellow, IEEE},
and~Qi~Bi,\IEEEmembership{Fellow, IEEE}

\thanks{A version of the manuscript was submitted to IEEE Journal of Selected Topics in Signal Processing, special issue on Autonomous and Evolutive Optimization in Networked AI (AEON). Liang Song and Jiayue Jin are co-first authors, and are with the College of Intelligent Robotics and Advanced Manufacturing, Fudan University, Shanghai 200433, China. They are also with the International Networking Systems of Artificial Intelligence Limited, Macao. Corresponding author: Liang Song  (e-mail: songl@fudan.edu.cn).}
\thanks{Peng Sun is with the Division of Natural and Applied Sciences, Duke Kunshan University, Kunshan 215316, China (e-mail: peng.sun568@duke.edu).}
\thanks{Di Li is with the Faculty of Electrical Engineering and Computer Science, Ningbo University, Ningbo 315000, China (e-mail: lidi1@nbu.edu.cn).}
\thanks{Yang Liu is with the College of Electronic and Information Engineering, Tongji University, Shanghai 201804, China (e-mail: yangliu25@tongji.edu.cn).}
\thanks{Athanassios Skodras is with the Department of Electrical and Computer Engineering, University of Patras, Patras 26504, Greece (e-mail: skodras@upatras.gr).}
\thanks{Dimitris Ampeliotis is with the Department of Digital Media and Communication, Ionian University, Argostoli 28100, Greece (e-mail: ampeliotis@ionio.gr).}
\thanks{Amit Dvir is with the Department of Computer Science, Ariel University, Ariel 40700, Israel
(e-mail: amitdv@g.ariel.ac.il).}
\thanks{Jiangchuan Liu is with the School of Computing Science, Simon Fraser University, Burnaby, BC V5A 1S6, Canada (e-mail: jcliu@sfu.ca).}
\thanks{V. C. M. Leung is with the Department of Electrical and Computer Engineering, The University of British Columbia, Vancouver, BC V6T 1Z4, Canada (e-mail: vleung@ieee.org).}
\thanks{Qi Bi and Yong Yuan are with the China Telecom Research Institute, Beijing 102218, China (e-mail: qibi@chinatelecom.cn).}
}

\begin{document}

\maketitle

\begin{abstract}

Recent advances in Artificial Intelligence (AI), particularly large-scale deep learning, have been driven by monolithic or offline training on massive curated datasets. While successful, these approaches face structural limitations: escalating data and computational demands, limited online adaptability, and a lack of system-level intelligence in open, non-stationary environments. This paper presents Networked AI — a networked intelligent system composed of interacting, heterogeneous, and continuously evolving AI agents. We establish a hierarchical layered architecture comprising an Agent Layer for heterogeneous capability construction, an Interaction Layer for signal-processing-guided online data extraction, and a Regulation Layer for stability-constrained collective evolution. Foundations in data science, signal processing, and system control are hereby synthesized into a unified theoretical framework of Networked AI, providing superior data efficiency, intelligence efficiency, interpretability, and scalability. Theoretical and experimental  analysis demonstrates that Networked AI can approach near-linear scaling in resource demand, in contrast to the exponential increase in standalone end-to-end systems. It provides not only online adaptation to out-of-distribution environmental novelties, but also the autonomous evolution towards application-specific intelligence.  Application frontiers are discussed for IoT spaces, embodied, spatial, and industrial intelligences, as well as the cellular infrastructure and beyond. Key open challenges are identified towards realizing locally autonomous, collectively intelligent, continuously evolving, globally regulated Networked AI. 
\end{abstract}


\section{Introduction}

In the past decade, with the introduction of key technologies such as AlexNet~\cite{krizhevsky2012imagenet}, ResNet~\cite{he2016deep}, Transformer~\cite{NIPS2017_3f5ee243}, etc., Artificial Intelligence (AI) has entered a new era driven by foundation models, large-scale self-supervised learning, and transformer-based architectures. Recent advances in large language models (LLMs)~\cite{GPT-1}, multimodal systems~\cite{CLIP}, and generative AI~\cite{Zero-Shot} have demonstrated unprecedented capabilities in language understanding, reasoning, perception, and content generation. At the same time, the rapidly growing applications of industrial intelligence, robotics, autonomous systems, information-physical infrastructure and embedded artificial intelligence increasingly require artificial intelligence systems to operate in larger-scale, dynamic and open environments, and put forward higher requirements for adaptability, efficiency, reliability and interpretability.

However, the current artificial intelligence system is still basically based on an independent end-to-end model, and intelligence is mainly obtained by offline training of a large number of static data sets, and then inferred from the deployment time. As the model scale continues to grow, there are more than trillions of parameters (for example, OpenAI GPT-3 with 175B~\cite{GPT_3}, Google PaLM with 540B~\cite{PaLM}, 405B Meta Llama 3.1 of \cite{grattafiori2024llama}), the independent end-to-end AI system has encountered unprecedented bottlenecks in data availability, computing scalability, online adaptability and system security. Recent scale law research~\cite{Scaling_laws} shows that the growth of model capabilities is increasingly dependent on exponential growth calculations and high-quality training data, and the availability of such data is becoming a major limiting factor. In addition, static pre-training models often find it difficult to continuously adapt to changing environments, long-tail scenarios and interactive real tasks after deployment.

Technically speaking, major artificial intelligence researchers in academia and industry pay more and more attention to these problems. In~\cite{TheBitterLesson}, Richard Sutton presented that scalable intelligence ultimately should come from systems capable of continual learning and adaptation rather than handcrafted priors alone. Further, LeCun \textit{et al.}~\cite{NEURIPS2025_LeCun2,NEURIPS2025_LeCun1} also repeatedly pointed out that current auto-regressive language models are insufficient for achieving human-level intelligence due to the fundamental limitations and constraints of language, as they lack persistent world models, planning, memory, and interaction with the physical world. In other words, understanding and interacting with the physical world are essential to achieving human-level intelligence. Similarly, Feifei Li \textit{et al.}~\cite{Yang_2025_CVPR} also emphasized that the next generation of AI must move beyond static language intelligence toward embodied and spatial intelligence grounded in interaction with the real world. Furthermore, Geoffrey Hinton and Yoshua Bengio, respectively, have recently raised concerns about the risks and limitations of purely large-scale black-box optimization, which is the foundation of the current AI paradigm, particularly in terms of controllability and interpretability. Bengio \textit{et al.} in~\cite{bengio2025superintelligentagent} further introduced the Scientist AI and trustworthy AI systems capable of causal reasoning, uncertainty awareness, and safety guarantees, rather than relying solely on predictive scaling. All these emerging viewpoints collectively suggest that AI is approaching a paradigm transition, i.e., from isolated model-centric intelligence toward continuously evolving system-level intelligence. The key challenge is that we must no longer focus solely on building larger models but shift our focus to constructing scalable AI systems capable of continuously generating high-quality adaptive data, collaboratively organizing intelligence, and safely evolving in open environments.

Accordingly, in this paper, to the best of our knowledge, we firstly introduce and comprehensively elaborate a new AI paradigm named Networked Artificial Intelligence (Networked AI), which represents a promising next-generation AI paradigm aimed at addressing the challenges mentioned above. \textit{Networked AI refers to a networked intelligent system where heterogeneous AI agents can continuously generate high-quality data through mutual interactions, perform autonomous and evolutive optimization of their behaviors in real time, adapt to new environmental conditions, and maintain system-level stability, safety, and trustworthiness. }As the application-tier realization of Networking System of AI (NSAI)~\cite{nsaisong}, Networked AI can significantly reduce AI systems’ dependence on massive centralized datasets and exponentially growing compute resources, while continually improving application-specific intelligence in continuous, open real-world scenarios.

Compared to traditional standalone end-to-end AI, Networked AI introduces two critical mechanisms absent in current large-model paradigms, i.e., \textit{interaction and regulation}. Networked AI dynamically extracts high-quality data and generates reliable pseudo-labels or optimization signals during operation~\cite{qian2024new,qian2026coordinated} by exploiting adaptive interactions guided by prior knowledge, policies, and environmental feedback. Meanwhile, local agent evolution and global system stability can be jointly optimized by leveraging hierarchical regulation mechanisms~\cite{bengio2026internationalaisafetyreport,Bengio.Hinton.AI_Risks}. These mechanisms can fundamentally improve the following unique features.
\begin{itemize}
    \item \textbf{Data efficiency}: In Networked AI, traditional static large-scale offline datasets will be replaced by continuously generated high-quality interactive data.
    \item \textbf{Intelligence efficiency}: Networked AI can carry out online evolution and continuous optimization in a dynamic environment, not just relying on static pre-training.
    \item \textbf{Interpretability}: Networked AI can further transform current AI from black-box optimization to partially structured and explainable gray-box systems through explicit organizational architectures and controllable interactions.
    \item \textbf{Scalability}: By introducing hierarchical agent organizations capable of decomposing and collaboratively solving arbitrarily complex tasks, Networked AI can overcome the context-window and sequential reasoning limitations of existing end-to-end models.
\end{itemize}

\begin{figure*}[!t]
\centering
\includegraphics[width=6in]{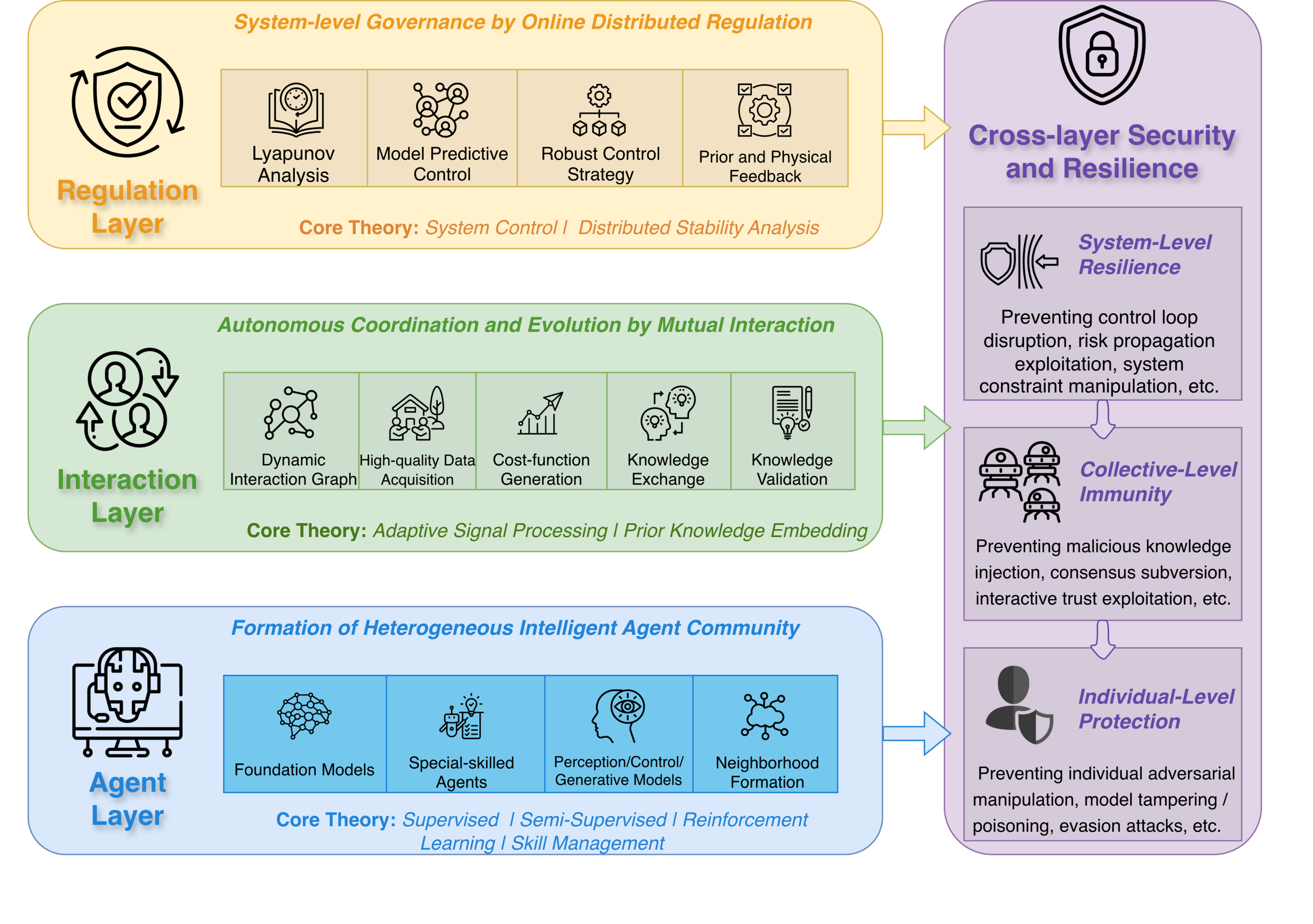} 
\caption{Layered architecture of Networked AI.}
\label{fig:architecture}
\end{figure*}

In Networked AI, the source of intelligence shifts from conventional individual model generalization, to collective and autonomous evolution towards application-specific scenarios. It can  present an architectural and system-level embodiment of Online Evolutive Learning (OEL)~\cite{qian2026coordinated,jin2024oel,oeldl}, which deals with the online extraction of high-quality data and the evolutive adaptation of heterogeneous models at the algorithmic level. Technically, traditional standalone AI often exhibits exponential growth in data and computational resource demands as system and task complexities increase. In contrast, Networked AI enables collaborative reuse of knowledge, distributed online optimization, and localized autonomous adaptation, which in turn shifts scaling behavior towards linear growth regimes. It thereby can transform the scaling relationship between system capability and resource demands. In this paper, we discuss the emerging architecture, theory and applications of Networked AI, as well as its industrial impacts and open challenges. Our main contributions are summarized as follows:
\begin{itemize}
    \item We establish a hierarchical layered architecture for Networked AI (Fig.~\ref{fig:architecture}), enabling collaborative optimization among large-scale AI agents. Specifically, in the bottom agent layer, we provide scalable heterogeneous agent communities based on data-driven models; in the middle interaction layer, we discuss online extraction of high-quality data through adaptive signal processing guided by prior knowledge and policies; in the top regulation layer, we introduce local-global regulation and control mechanisms maintaining system stability during continual evolution.
    \item We analyze the scaling relationships between intelligence capability, data and computation demand, and the number of heterogeneous task components, in Networked AI versus standalone end-to-end AI, highlighting the near-linear versus exponential scaling trends; and we also discuss resilient cross-layer security provided by the interpretable layered architecture.
    \item We further review representative applications in embodied, spatial, and industrial AI, as well as IoT and cellular infrastructures, emphasizing potential transformative gains in the application frontiers.
    \item Finally, we identify open research challenges in Networked AI, which represent a new scientific paradigm bridging supervised learning, reinforcement learning, continual learning, and distributed intelligence into the next-generation AI systems.
\end{itemize}

Rather than viewing large foundation models as the endpoint of AI evolution, we argue that those existing large models may instead become foundational components within larger networked intelligent ecosystems. The future of AI may therefore depend less on building even-larger isolated models, but more on constructing continuously evolving, trustworthy, and collaboratively intelligent networked systems. Accordingly, these newly introduced capabilities will establish a novel closed-loop AI paradigm integrating system architecture, learning, regulation, security, and verification.

\section{ System Framework}



%


\subsection{Definitions}

\label{se:sysabs}
A Networked AI system can be formally abstracted as a time-evolving, spatially distributed dynamical system that fundamentally comprises a universal set of heterogeneous intelligent entities, denoted as $\mathcal{A} = \{a_1, a_2, \dots, a_N\}$. Each agent $a_i$ is an intelligent unit providing a relatively independent function within the overall system. To accommodate disparate architectures and learning paradigms within a common framework, an agent is formally represented as a tuple:
\begin{equation}
    a_i=(\mathcal{X}_i, \mathcal{Y}_i, \Theta_i, \Pi_i), \label{Eq: agent representation}
\end{equation}
where $\mathcal{X}_i$ and $\mathcal{Y}_i$ denote the input and output spaces, respectively. $\Theta_i$ represents the model parameters or rule set. $\Pi_i$ denotes the specific update mechanism.
This abstraction establishes a unified systems view for incorporating diverse functional models across the network. These agents do not operate in isolation; rather, the entire networked system is embedded within a non-stationary external environment $\mathcal{E}(t)$. The spatial interaction topology among these agents at any given time $t$ is represented by a dynamically shifting graph $G(t)$. 

Unlike standalone models that map static datasets to fixed parameter spaces, the Networked AI system maintains a distributed global dynamic state $S(t)$, which encapsulates the real-time cognitive and operational conditions of all agents across the network. Driven by networked interactions and bounded by regulation, the macro-level dynamic evolution of the entire system can be abstracted by a generalized global transition operator $\boldsymbol{\mathcal{H}}$, which is given by,
\begin{equation}
\label{eq:global_update}
    S(t+1) = \boldsymbol{\mathcal{H}}\Big(S(t), \boldsymbol{\mathcal{Z}}(t), \mathcal{E}(t), \boldsymbol{u}(t)\Big),
\end{equation}
where $\boldsymbol{\mathcal{Z}}(t)$ represents the global integrated interaction flow across the dynamic graph $G(t)$ and $\boldsymbol{u}(t)$ encapsulates the top-down global regulatory control inputs.
However, resolving the global operator $\boldsymbol{\mathcal{H}}$ directly is computationally intractable due to the massive scale and extreme heterogeneity of the agents. To make this system practically computable and scalable, the macro-level transition is decoupled into localized, autonomous state transitions. Specifically, for an arbitrary agent $a_i$, its local state transition is governed by its internal dynamic state $s_i(t)$ which inherently includes its evolvable parameters $\Theta_i(t)$:
\begin{equation}
\label{eq:local_update_abstract}
    s_{i}(t+1)=\mathcal{H}_{i}\left(s_{i}(t), \mathcal{Z}_{i}(t), \mathcal{E}(t), u_{i}(t)\right),
\end{equation}
where $\mathcal{H}_i(\cdot)$ is the internal update mechanism, $\mathcal{Z}_i(t)$ is the localized interaction flow filtered from neighboring peers, and $u_i(t)$ is the localized stability-oriented control input.

Instead of minimizing a static empirical risk, the fundamental objective of Networked AI is to continuously drive the global state $S(t)$ towards optimal functional consensus and system-wide stability while adapting to the non-stationary environment $\mathcal{E}(t)$. To achieve this, the system relies on a hierarchical coordination framework composed of two complementary mechanisms, i.e., 1) a bottom-up interactive learning process: during localized interactions, agents continuously extract high-quality data and formulate evolutive learning objectives by processing information exchanged with their peers. And 2) a top-down regulation process: the control input $u_i(t)$ provides top-down regulation to constrain local updates and ensure global stability.
By seamlessly coupling interaction-driven data extraction with macro-level regulation, the system ensures that localized autonomous evolution stably and robustly serves the collective intelligence without diverging.

To ensure mathematical clarity and consistency throughout the theoretical formulation, a few symbols used in this paper are summarized in Table~\ref{tab:nomenclature}. The subscript $i$ denotes an individual agent. $(t)$ denotes the discrete time step.

\begin{table}[htbp]
\centering
\caption{Nomenclature of the Networked AI Framework}
\label{tab:nomenclature}
\renewcommand{\arraystretch}{1.3}
\begin{tabular}{l p{7cm}}
\toprule
\textbf{Symbol} & \textbf{Definition / Description} \\
\midrule
$\mathcal{A}$ & Universal set of all heterogeneous intelligent entities \\
$N$           & Total number of agents in the network \\
$a$           & An individual autonomous intelligent agent \\
$\mathcal{X}, \mathcal{Y}$ & Input and output spaces of an agent (respectively) \\
$\Theta$      & Internal model parameters or rule set of an agent \\
$\Pi$         & Specific update mechanism / learning paradigm of an agent \\
$t$           & Discrete time step index \\
$\mathcal{E}$ & Non-stationary external environment \\
$G$           & Dynamically shifting spatial interaction graph / topology \\
$S, s$        & Global dynamic state of the system, and local dynamic state of an agent \\
$\boldsymbol{\mathcal{H}}, \mathcal{H}$ & Global transition operator, and local internal update mechanism \\
$\boldsymbol{\mathcal{Z}}, \mathcal{Z}$ & Global integrated interaction flow, and localized interaction flow \\
$\boldsymbol{u}, u$ & Global regulatory control inputs, and local stability-oriented control input \\
\bottomrule
\end{tabular}
\end{table}

\subsection{Layered Architecture}

In order to systematically describe the functional organization of Networked AI, we adopt a three-layer architecture that separates capability building, local evolution and system-level regulation:

\begin{enumerate}
\item \textbf{Agent Layer:} It provides the basic capability unit of the system, which is composed of heterogeneous agents with different model structures, learning paradigms and functional specialization. It focuses on local adaptability under representation learning, task-specific modeling and resource constraints.

\item \textbf{Interaction Layer:} It controls the decentralized coordination and local evolution process between agents. It defines the interactive topology $G(t)$ and realizes evolutive adaptation through neighborhood communication, consistency enforcement and other mechanisms. The interaction layer is responsible for converting local updates into collective evolution for specific application systems.

\item \textbf{Regulation Layer:} It regulates the behavior of the entire network system by introducing global objectives, a priori constraints and feedback signals, thus providing system-level control. Key functions include stability control, convergence adjustment and reducing error propagation in large-scale distributed applications.
\end{enumerate}

The three layers jointly define a hierarchical structure in which local intelligence, interaction dynamics, and global regulation are coherently integrated. This abstraction distinguishes Networked AI from traditional paradigms such as federated learning, multi-agent reinforcement learning, and swarm intelligence, by explicitly incorporating communication-based continual evolution and system-level control into an autonomous intelligence system.

\section{Theoretical Foundations}

\subsection{Agent Layer: Formation of Heterogeneous Intelligent Agent Community}
The Agent Layer provides the fundamental capability units of Networked AI to learn from online data. The system integrates a diverse set of intelligent models with different functional roles and learning capacities instead of assuming homogeneous agents. Agents include foundation models, such as large language models (LLMs)~\cite{radford2019language,grattafiori2024llama,guo2025deepseek,yang2025qwen3}, vision-language models (VLMs)~\cite{liu2023visual,team2023gemini,openai2024gpt4o,bai2025qwen3vl}, and vision-language-action (VLA) models~\cite{kim2024openvla,black2024pi_0,zhou2026opendrivevla,zuo2026embodied}; skill-oriented models which are used for planning, tool utilization, and workflow execution~\cite{yao2022react,schick2023toolformer,wu2024autogen}; traditional perception modules such as Convolutional Neural Networks~\cite{krizhevsky2012imagenet,simonyan2014very,he2016deep,howard2017mobilenets,liu2023amp,chen2026cross} and Vision Transformers~\cite{dosovitskiy2020image,touvron2021training}); conventional decision-making modules like Markov Decision Processes~\cite{puterman2014markov} and reinforcement learning agents~\cite{sutton1998reinforcement}); and lightweight edge AI models operating under latency, memory, and energy constraints~\cite{tan2019mnasnet,lin2020mcunet}. This heterogeneity allows the system to combine general reasoning abilities with domain-specific skills, forming a rich capability space. 

 A primary challenge lies in their uniform management and representation since these models exhibit diverse functions, architectures, and update mechanisms. As formally abstracted in Section~\ref{se:sysabs}, Networked AI addresses this by formalizing each functional model as an autonomous agent tuple $a_i=(\mathcal{X}_i,\mathcal{Y}_i,\Theta_i,\Pi_i)$ within the universal set $\mathcal{A}$ (see Eq.~\eqref{Eq: agent representation}). Building upon this formalization, the Agent Layer mainly focuses on organizing these discrete entities into a structured and scalable topology.

When the number of agents becomes large, a flat organization quickly becomes intractable. Placing all agents in a single conceptual plane leads to profound coordination complexity and diminished scalability. To address this, Networked AI employs a scalable, hierarchical agent architecture. Let the Agent Layer consist of $L$ levels,
\begin{equation}
\mathcal{A}=\bigcup_{l=1}^{L}\mathcal{A}^{(l)}, \qquad
\mathcal{A}^{(l)}=\left\{a^{(l)}_{1}, a^{(l)}_{2}, \dots, a^{(l)}_{K_l}\right\},
\label{eq:hierarchy}
\end{equation}
where $a^{(l)}_{i}$ denotes the $i$-th agent at level $l$, and $K_l$ is the number of agents at that level. For $l \geq 2$, each higher-level agent is defined as a collection of lower-level agents, representing a functional abstraction,
\begin{equation}
a^{(l)}_{i}=\left\{a^{(l-1)}_{j} \mid j \in \Omega^{(l)}_{i}\right\},
\label{eq:containment}
\end{equation}
where $\Omega^{(l)}_{i}$ is the index set of child agents contained within $a^{(l)}_{i}$. 

Depending on their respective functional roles, individual agents within this hierarchy may be optimized using diverse paradigms (e.g., supervised learning, reinforcement learning, or explicit skill construction). Regardless of the specific learning paradigm employed, their evolution dynamics follow the unified state update formulation introduced in Eq.~\eqref{eq:local_update_abstract}. Specifically, situated within this hierarchical architecture, the dynamic state evolution of an agent $a_{i}^{(l)}$ at level $l$ is explicitly contextualized as,
\begin{equation}
s_{i}^{(l)}(t+1)=\mathcal{H}_{i}^{(l)}\left(s_{i}^{(l)}(t), \mathcal{Z}_{i}^{(l)}(t), \mathcal{E}(t), u_{i}^{(l)}(t)\right),
\label{eq:state_update}
\end{equation}
where the general operator $\mathcal{H}_i^{(l)}(\cdot)$ governs the internal update, the crucial structural feature introduced by the Agent Layer is the integration of cross-level coupling. In this hierarchy, the localized interaction flow $\mathcal{Z}_{i}^{(l)}(t)$ dynamically fuses both horizontal peer interactions and vertical functional abstractions through a macro-operator $\mathcal{F}(\cdot)$,
\begin{align}
\mathcal{Z}_{i}^{(l)}(t) = \mathcal{F} \left( \sum_{k}\alpha_{ik}^{(l)}s_{k}^{(l)}(t), \;
 \mathcal{I}^{(l)} \left[ \sum_{j}\beta^{(l-1\rightarrow l)}_{ij}s^{(l-1)}_{j}(t) \right] \right),
\label{eq:interaction_flow}
\end{align}
where $\sum_{k}\alpha_{ik}^{(l)}s_{k}^{(l)}(t)$ captures the horizontal peer interaction with neighboring agents at the same level $l$ and $\mathcal{I}^{(l)} \left[ \sum_{j}\beta^{(l-1\rightarrow l)}_{ij}s^{(l-1)}_{j}(t) \right]$ denotes the vertical functional abstraction from lower-layer child agents. $\alpha_{ik}^{(l)}$ represents the interaction strength between peer agents. $\beta^{(l-1\rightarrow l)}_{ij}$ quantifies the contribution of lower-layer agents to the upper-layer abstraction. Technically, this non-linear mapping function $\mathcal{I}^{(l)}(\cdot)$ is strictly deployed to align, compress, and project disparate, low-level multi-modal representations (e.g., pixel-level detections) into the homogeneous algebraic space of the current level $l$. This formalized abstraction establishes a blueprint for the operational mechanics of the system. In the subsequent sections, the Interaction Layer (Section~\ref{se:interaction}) explicitly instantiates the macro-operator $\mathcal{F}(\cdot)$ by detailing its constituent steps of signal processing and cost formulation, mapping the abstract interaction flow $\mathcal{Z}_{i}^{(l)}(t)$ to its operational data-loss pair $(\mathcal{D}_i^*(t), J_i(t))$. Meanwhile, the Regulation Layer (Section~\ref{se:regulation}) operationally governs the state transition function $\mathcal{H}_{i}^{(l)}(\cdot)$ by injecting the stability-oriented control input $u_{i}^{(l)}(t)$ derived from Lyapunov theory, ensuring that the localized parameter evolution converges robustly toward the global system objective.





\subsection{Interaction Layer: Autonomous Evolution and Coordination by Mutual Interaction}
\label{se:interaction}

As mentioned above, the Agent Layer provides heterogeneous capabilities, while the Interaction Layer governs how agents undergo local autonomous evolution through adaptive signal processing. This layer essentially defines the knowledge and message exchange mechanisms between agents, ensuring that all interactions are explainable and grounded in prior knowledge, rather than functioning in a black-box manner.
Existing signal processing theories/methods provide the theoretical foundation for this interpretable exchange. For instance, optimal estimation techniques such as Kalman filter~\cite{kalman1961new} and Wiener filtering~\cite{wiener1949extrapolation} are employed to denoise inter-agent signals, while adaptive algorithms like Least Mean Squares (LMS) and Recursive Least Squares (RLS)~\cite{plackett1950some} allow agents to adjust to non-stationary environments. Furthermore, Graph Signal Processing (GSP)~\cite{shuman2013emerging} is utilized to analyze information flow across the interaction graph $G(t)$, and Bayesian inference~\cite{kingma2013auto} enables the fusion of signals with application-specific priors. These methods act as functional abstractions that ensure every piece of exchanged information can be analyzed and verified based on the system's prior knowledge, transforming raw interactions into a structured dialogue. Recent studies have further shown that structural knowledge of sensing can be explicitly embedded into online optimization, improving both sample efficiency and explainability under dynamically changing conditions \cite{pulkkinen2024model}. This observation supports the Interaction-Layer design principle that prior-guided adaptive signal processing can provide an interpretable foundation for evolution.

An essential function of the Interaction Layer is to continuously extract high-quality data and generate a corresponding cost function for every interactive agent. Mathematically, to instantiate the interaction flow defined in Eq.~\eqref{eq:interaction_flow}, an agent continuously utilizes the aforementioned signal processing tools to filter the raw information exchanged with its active neighbors over the dynamic graph $G(t)$. This filtering process yields a reliable, denoised network-derived actual signal, denoted as $y_i(t, G(t)) \in \mathcal{Y}_i$, effectively grounding the interaction within the agent's predefined output space. The explicit notation of both $t$ and $G(t)$ formalizes a critical spatio-temporal coupling, i.e., it emphasizes that this consensus signal depends simultaneously on the continuous temporal evolution over time $t$ of the incoming data streams and the structural constraints imposed by the dynamically shifting network topology $G(t)$. Concurrently, the agent interacts with the non-stationary environment $\mathcal{E}(t)$ and monitors the residual error between its own internal prediction $\hat{y}_i(t) \in \mathcal{Y}_i$ and this interactive observation. A statistical shift or unfamiliar scenario induced by the evolving environment $\mathcal{E}(t)$ is detected when this discrepancy exceeds a predefined threshold $\tau$, i.e., $||\hat{y}_i(t) - y_i(t, G(t))|| > \tau$. Triggered by this environment-induced discrepancy, the agent initiates a targeted adaptation process. Using a localized extraction function $\Psi$, it processes this error to identify a set of high-quality data $\mathcal{D}_i^*$ and formulate a specific local cost function $J_i$. We formalize this interaction-driven extraction process as follows,
\begin{equation}\label{eq:eq8}
(\mathcal{D}_i^*(t), J_i(t)) = \Psi(\hat{y}_i(t), y_i(t, G(t)), \mathcal{E}(t), \tau).
\end{equation}
This abstraction ensures that the system selectively filters interaction signals over $G(t)$ to isolate high-fidelity components that reflect the current state of $\mathcal{E}(t)$. For example,  a detailed illustration of interaction-driven joint optimization process is given in~\cite{qian2026coordinated}.

This interaction-driven update mechanism enables the system in transition from offline generalization to online evolutive adaptation. By executing targeted data extraction and parameter updates solely upon detecting statistical novelties in the signal space, as well as the prior knowledge embedded in the adaptive signal processing for agent interactions, the Interaction Layer supports localized feature processing and avoids unrestricted system-wide Cartesian expansion, providing the structural basis for the near-linear scaling regime.




\subsection{Regulation Layer: System-Level Governance by Online Distributed Regulation}
\label{se:regulation}

The Regulation Layer establishes global mechanisms to coordinate collective behavior and ensure long-term stability within the Networked AI architecture. The theoretical foundations for this layer are rooted in theory of system control, particularly Lyapunov stability analysis~\cite{lyapunov1992general}, Model Predictive Control~\cite{richalet1978model}, and robust control strategies. Lyapunov method provides a mathematical framework for proving the convergence of dynamic systems by constructing a scalar energy function, while robust control solves the problem of performance margin under environmental uncertainty. In Networked AI, these theories allow a formal assessment of the evolutive path of heterogeneous agent communities and prevent system-wide differences.


Traditional data-driven models are often known as black-box systems. Due to the absence of interpretable input-output mappings, it is very difficult to apply system control methodologies directly to the training process. However, the proposed Networked AI addresses this limitation by utilizing the Interaction Layer to convert raw signals into interpretable relationships grounded in prior knowledge. This structural transformation effectively enables the intelligent system to be treated as a gray-box that can be formally described and regulated. The Regulation Layer can then mediate the relationship between local autonomous evolution and global system requirements, by incorporating physical feedback and high-level constraints into the governance cycle.

We hereby introduce a scalar Lyapunov candidate function $V(S(t))$ representing the global energy or collective error of the overall system. To align with the hierarchical architecture defined in the Agent Layer, the global state $S(t)$ is formulated as the universal set of all dynamic agent states across the network, i.e., $S(t) = \left\{ s_{i}^{(l)}(t) \mid 1 \le l \le L, 1 \le i \le K_l \right\}$, where $s_{i}^{(l)}(t)$ is introduced in Eq.~\eqref{eq:state_update}. Specifically, $V(S(t))$ acts as the dynamic evaluation metric for the entire system. As the collective minimization of local costs $J_i$ drives the descent of the global summation term, the system-level constraints are encoded within the energy landscape of $V(S(t))$. However, continuously computing the exact global variation requires centralized aggregation, which can be computationally prohibitive and may violate decentralized constraints of some Networked-AI applications. To resolve this, the Regulation Layer may assume that the global energy function is additively separable or can be locally approximated, i.e., 
\begin{align}\label{eq:eq9}
V(S(t)) \approx \sum_{i} V_i(s_i(t), \mathcal{Z}_i(t)),\quad 1\leq i\leq K.    
\end{align}
Consequently, each agent evaluates a predicted local-energy variation under the unregulated candidate update, denoted as $\Delta V_i(t)$, which estimates its individual contribution to the global system dissipation based solely on its state and neighborhood interaction flow. To approach asymptotic stability, the stability-oriented control input $u_{i}^{(l)}(t)$ in Eq.~\eqref{eq:state_update} is operationally instantiated as a local regulation operator, i.e., $u_{i}^{(l)}(t) := \Gamma_{i}(t)$. This operator is formulated to set up local passivity, which can collectively achieve system-wide dissipation:
\begin{equation}\label{eq:eq10}
    \Gamma_i(t) = 
    \begin{cases} 
        1, & \text{if } \Delta V_i(t) \le 0 \\
        \exp(-\kappa \Delta V_i(t)), & \text{if } \Delta V_i(t) > 0 
    \end{cases}
\end{equation}
where $\kappa$ is a predefined positive scaling coefficient. For an agent parameterized by $\Theta_i$ (omitting the layer superscript $(l)$ for notational simplicity), the regulated parameter update is subsequently expressed as follows:
\begin{equation}\label{eq:eq11}
\Theta_i(t+1) = \Theta_i(t) - \Gamma_i(t) \cdot \eta \nabla_{\Theta_i} J_i(\mathcal{D}_i^*, \Theta_i),
\end{equation}
where $\Gamma_{i}(t) \in (0,1]$ adjusts the effective learning rate $\eta$. This regulated update law provides a concrete and parametric realization of the abstract state transition function $\mathcal{H}_{i}^{(l)}(\cdot)$ by treating the model parameters as the core evolvable component of the agent's dynamic state, i.e., $\Theta_i(t) \subseteq s_i^{(l)}(t)$. Specifically, the internal state transition is driven by the local cost $J_i$ and the extracted data $\mathcal{D}_i^*$, which directly operationalize the abstract interaction flow $\mathcal{Z}_{i}^{(l)}(t)$. This implicitly encodes the influence of horizontal peer interaction and vertical function abstraction from the local neighborhood of the agent. It in principle ensures that the localized, heterogeneous parameter evolution $\Theta_i$ may converge toward the global system objective by incorporating the environmental feedback through $\mathcal{D}_i^*$ and coupling it with the Lyapunov regulation factor $\Gamma_i(t)$, thereby maintaining the robustness of the collective intelligence in non-stationary settings (Theorem \ref{thm:stability}, Appendix \ref{app:stability}).

\subsection{Efficiency of Collective Intelligence}

While the preceding sections have established the theoretical foundation of the proposed three-layer Networked AI architecture, we hereby further connect the layered architectural design to its complexity scaling analysis from both theoretical and experimental perspectives.  


From a theoretical perspective, a standalone end-to-end system needs to jointly process heterogeneous tasks, as comparable to those from all $N$ agents in Networked AI. \textit{Therefore, $N$ can virtually indicate the scale of overall task complexity or system capability for both systems. }This global coupling leads to an exponential training-data requirement of $\Omega(c^N)$ for standalone systems (Theorem \ref{thm:standalone_scaling}, Appendix \ref{app:scaling_theorems}), where $c$ denotes the independent feature complexity from an individual task. 

In contrast, the hierarchical multi-agent community, as specified in the Networked-AI architecture and theory, is able to structurally reduce the system coordination depth, effectively lowering the overall task degree of freedom from $O(N)$ as in standalone systems, to $O(\log N)$ in Networked AI, where $\log N$ is the number of hierarchical levels. Therefore, the corresponding training data requirement can be reduced to $O(N \cdot c^{\log N})$, which can actually be close to a near-linear scaling of $N$ in Networked AI (Theorem \ref{thm:networked_scaling}, Appendix \ref{app:scaling_theorems}). The reduction is dependent on the system design with embedded depth and prior knowledge in the Networked-AI architecture.

The above theoretical views give a scaling comparison: when the overall task complexity or system capacity increases, the resources required by standalone end-to-end system incurs superlinear or exponential growth, and Networked AI can approach near-linear growth through localized agents decomposition learning and hierarchical coordination. In order to verify the theoretic comparison, we have built a controlled experimental environment in which the model scale, application scale and overall task complexity can change independently. Under the matching task conditions, we compared the standalone AI configuration and the Networked AI configuration, and then fit the resource-scaling curve.

Because the Interaction Layer extracts task-relevant samples and only activates localized updates for deficient components, redundant data acquisition and global retraining are reduced. This on-demand and evolutive adaptation mechanism improves data and intelligence efficiency in a changing environment. Therefore, from the experimental perspective, we evaluated the computational cost of convergence under different models and application settings as an empirical proxy for interaction or data demand.

In the previous study, sensing-control decoupling and OEL experiments were introduced \cite{jin2025achieving}, and we extended the original local feedback loop to a hierarchical systems containing multiple heterogeneous agents. We employ a configurable grid-world in which agents navigate toward a target while avoiding known obstacles and dynamically introduced unknown obstacles. The following experiments include \textit{6 heterogeneous agents, 5 model scales, 8 application scenarios}, and the spatial size and obstacle density gradually increase. Eliminate unknown obstacles from the initial training distribution, simulate the non-stationary  environmental displacement $\mathcal{E}(t)$ and the novelty outside the distribution. In each time step, the agents receive a local observation, select a discrete movement action, and get a sparse rewards for penalizing collision and rewarding successful navigation. This experiment is mapped to the proposed three-layer architecture, as follows:
\begin{itemize}
\item \textbf{Agent Layer:} Assign heterogeneous perception and control functions to specialized agents. In the experimental implementation, these roles are instantiated by YOLOv5-based perception agents and DQN-based control agents. However, the architecture is not limited to these specific models and can accommodate other perception and decision-making modules.
\item \textbf{Interaction Layer:} Consistent with the theoretical formula, the discrepancy signal is defined as $\Delta_i=|R_p-R_a|$. Here, $R_p$ is the predicted reward, and $R_a$ is the actual reward, which correspond to $\hat{y}_i(t)$ and $y_i(t, G(t))$ in the interaction layer, respectively. When $\Delta_i>\tau$, the system extracts the corresponding interaction sample and stores it in the localized dataset $\mathcal{D}_i^*$.
\item \textbf{Regulation Layer:} The localized buffer triggers an asynchronous update $\Theta_i$ only for the deficient component, avoiding unnecessary global backpropagation and interference with unaffected modules.
\end{itemize}

\begin{figure}[t] 
    \centering 
    \subfloat[]{ 
        \includegraphics[width=0.97\columnwidth, trim=0 0 0 0, clip]{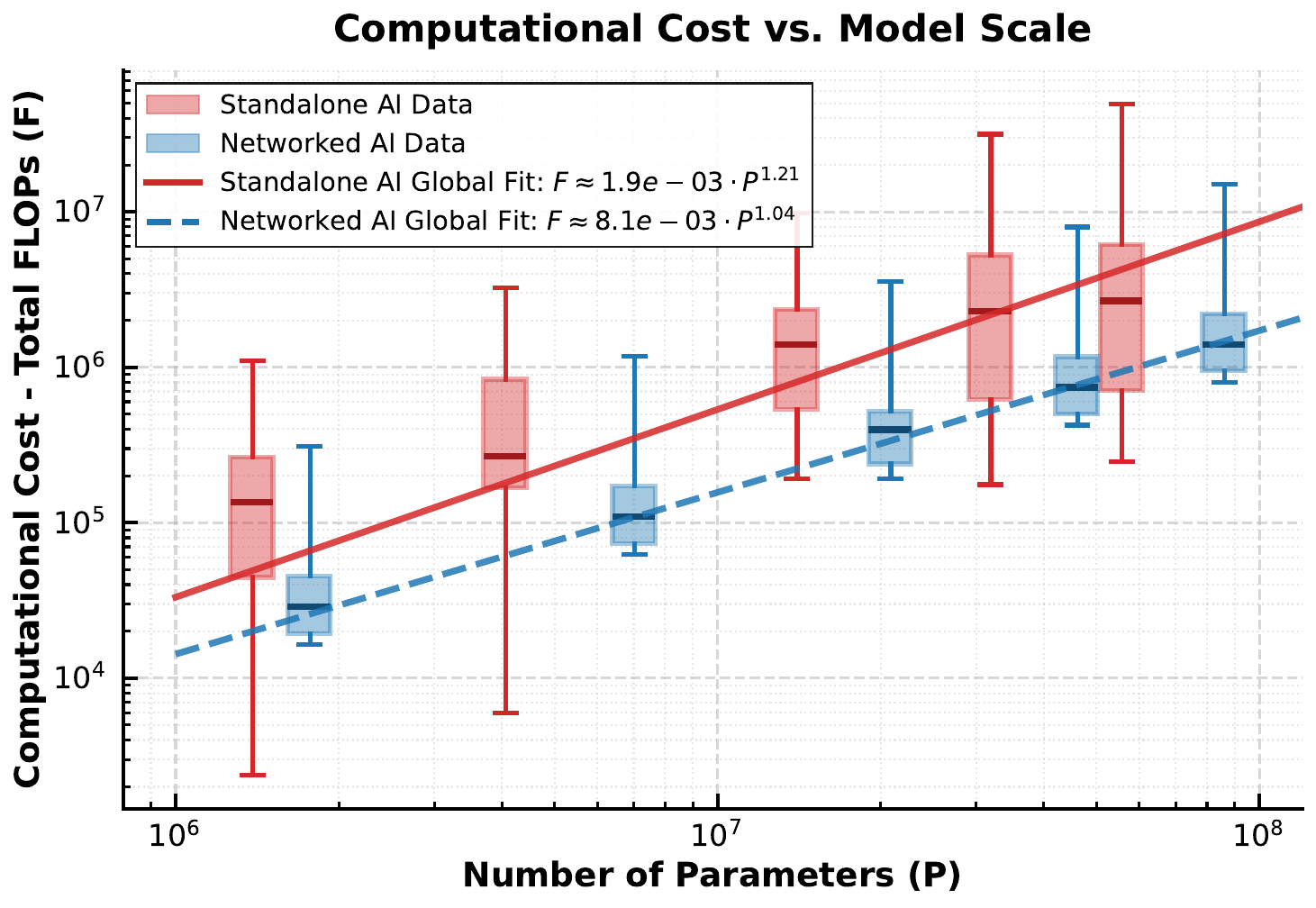} 
        \label{fig:model_scale_boxplot} 
    } 
    \vspace{-2mm}
    \subfloat[]{ 
        \includegraphics[width=0.97\columnwidth, trim=0 0 0 0, clip]{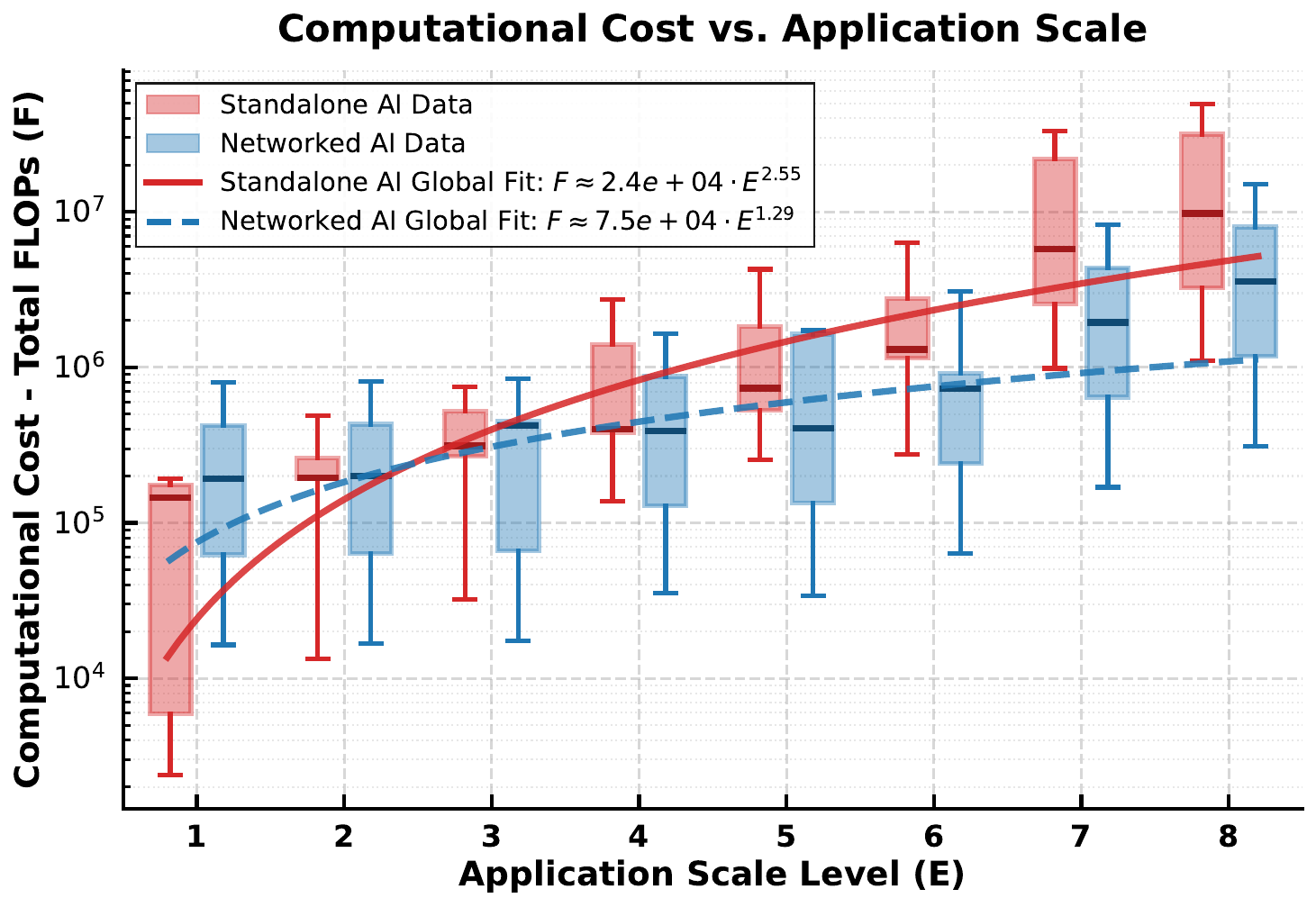} 
        \label{fig:task_complexity_boxplot} 
    } 
    \vspace{-2mm}
    \subfloat[]{ 
        \includegraphics[width=0.97\columnwidth, trim=0 0 0 0, clip]{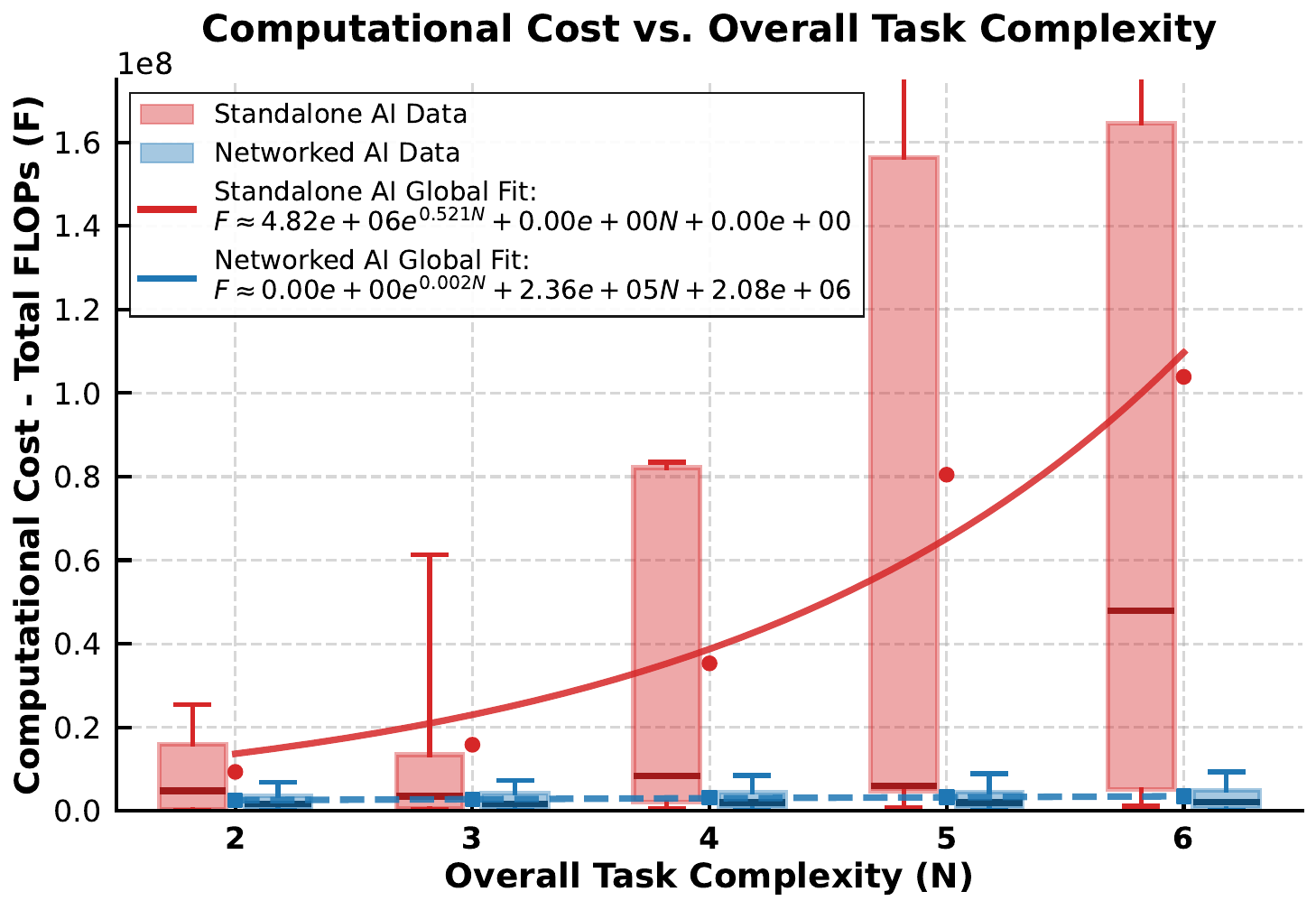} 
        \label{fig:task_agent_number_boxplot} 
    }
    \caption{Computational cost to convergence under different model, application and overall task complexity scales. The box plots summarize the variability across heterogeneous experimental configurations, while the solid and dashed curves represent the global power-law or exponential fits for standalone AI and Networked AI.} 
    \label{fig:complexity_scaling} 
\end{figure}

We measured the total FLOPs accumulated from experimental initialization to policy convergence. When an agent reaches the target, it does not collide with known or unknown obstacles within the predefined number of consecutive episodes, that is, convergence is achieved. Since the data sample is generated through the interaction between the agent and the environment, the convergent calculation cost can be used as an empirical proxy for interaction and data requirements.

For data visualization, the observations that share the same model scale, application scale or overall task complexity are grouped into a box plots, which summarize the median, quartiles and dispersion of different experimental configurations. The vertical axes for model and application scales are plotted logarithmically, whereas a linear scale is adopted for the overall task complexity (agent numbers). For model and application scales, the overall scaling trends are obtained by fitting the power-law model $F=kX^{\alpha}$ through log-log least-squares regression, where $X=P$ denotes the model parameter count and $X=E$ denotes the application scale level. For overall task complexity, represented by the agent number $N$, the arithmetic mean FLOPs at each $N$ are fitted using the exponential-linear model $F=ae^{bN}+cN+d$ through nonlinear least-squares optimization. The box plots characterize the variability in computational costs across heterogeneous experimental configurations, while the fitted curves capture the global resource-scaling behavior of standalone and Networked AI.

Fig.~\ref{fig:model_scale_boxplot} shows the relationship between model scale and computational cost. The horizontal axis represents the actual number of model parameters $P$. For each model scale, the box plot summarizes the distribution of convergence costs in different application scenarios, and the heterogeneous agent $N$ is only set to 2 and 3 to isolate the impact of the model scale. The global power-law fits are:
\begin{equation}
F_{\mathrm{standalone}}(P) \approx 1.9\times10^{-3}P^{1.21},
\end{equation}
and
\begin{equation}
F_{\mathrm{networked}}(P) \approx 8.1\times10^{-3}P^{1.04}.
\end{equation}

The exponent of standalone AI is super-linear, whereas Networked AI remains close to linear. This result indicates that increasing model capacity imposes a growing global re-optimization cost on the standalone system, while localized updates allow Networked AI to limit the additional interaction and data demand.

Fig.~\ref{fig:task_complexity_boxplot} depicts the effect of application scale, where the number of heterogeneous agents $N$ is still set at 2 and 3 only, to isolate the impact of application. For a grid world of size $H\times W$, we first calculate an application score
\begin{equation}
\Phi=(N_U+N_O)\sqrt{H\times W},
\end{equation}
where $N_U$ and $N_O$ denote the numbers of unknown objects and known obstacles, respectively. The evaluated scenarios are ordered according to $\Phi$ and mapped to the application-scale levels $E\in\{1,\ldots,8\}$. At each level, the box plot summarizes the convergence costs obtained under different model scales and agent configurations. The global power-law fits are
\begin{equation}
F_{\mathrm{standalone}}(E) \approx 2.4\times10^{4}E^{2.55},
\end{equation}
and
\begin{equation}
F_{\mathrm{networked}}(E) \approx 7.5\times10^{4}E^{1.29}.
\end{equation}
The substantially smaller exponent of Networked AI indicates that its training cost is less sensitive to increasing application scale.Since $N\in\{2,3\}$, the theoretical model predicts that globally coupled standalone learning may exhibit an effective dependence between $O(c^2)$ and $O(c^3)$, whereas Networked AI should exhibit a lower-order growth trend through functional decomposition and localized optimization. The empirical fitting, $E^{2.55}$ for standalone AI and $E^{1.29}$ for Networked AI, are matching with the theoretical base: the standalone result lies within the expected near-quadratic to near-cubic range, while the Networked AI shows substantially milder growth. 

Fig.~\ref{fig:task_agent_number_boxplot} represents computational cost as a function of the overall task complexity, indexed by the heterogeneous agent number $N\in\{2,3,4,5,6\}$. For the standalone systems, each $N$ value represents a comparable collection of heterogeneous tasks that must be processed jointly. For each $N$, the box plots summarize the convergence cost of five model scales and eight application scales. The fits are
\begin{equation}
 F_{\mathrm{standalone}}(N)
\approx 4.8224\times10^{6}e^{0.52077N},
\label{eq:standalone_overall_task_fit}
\end{equation}
and
\begin{equation}
 F_{\mathrm{networked}}(N)
\approx 2.3645\times10^{5}N+2.0763\times10^{6}.
\label{eq:networked_overall_task_fit}
\end{equation}
The fitted results reveal distinct computational scaling trends. Standalone AI exhibits exponential growth with an exponent of $0.52077$, whereas Networked AI follows an approximately linear trend, with an incremental cost of only $2.3645\times10^5$ FLOPs per additional agent. As $N$ increases from $2$ to $6$, the fitted computational cost increases by a factor of $F_{\mathrm{standalone}}(6)/F_{\mathrm{standalone}}(2)=e^{0.52077(6-2)}\approx8.03$ for standalone AI, compared with only $F_{\mathrm{networked}}(6)/F_{\mathrm{networked}}(2)=3.4950/2.5492\approx1.37$ for Networked AI.

Under the evaluated configurations, hierarchical decomposition and localized optimization in Networked AI substantially reduce the observed growth rate of data and computational demand. Together with Appendix \ref{app:scaling_theorems}, the results support a conversion of exponential-to-near-linear scaling, which can reduce the order of magnitude for data and computational resource demand in complicated tasks.

\subsection{Cross-layer and Resilient Security}

Networked AI transforms deep-learning based standalone AI from a traditional large black-box system into a gray-box where all interactions and regulations over the networked AI agents can be clearly interpreted. Following the principle that "security emerges from architecture", a multi-level defense method is required, including single agent reinforcement, collective anomaly recovery ability and system-level recovery. This involves the entire life cycle of protecting data, models and behaviors under dynamic and potentially unfavorable conditions. By integrating security into all layers of the architecture, Networked AI can achieve enhanced resilience, viability and robustness against known and unpredictable interruptions.

The transition from an independent end-to-end model to Networked AI has also introduced a different threat scenario. When an independent model presents a larger attack surface, Networked AI is composed of heterogeneous, interactive and evolving Agents, presenting a dynamic and multi-dimensional vulnerability space. For example, conventional threat modeling frameworks were designed for systems with static computational graphs, fixed data flows, and centralized control logic, which lack the expressiveness to model threats arising from agent autonomy, emergent inter-agent behavior, and the online adaptation mechanisms. Recent work such as MAESTRO~\cite{zambare2025securingagenticaithreat} represents an advance in agent-centric threat modeling, but is still scoped primarily to individual agents instead of the systemic, cross-layer threat dynamics inherent to Networked AI.

At the Agent Layer, each agent is represented by the tuple $a_i=(\mathcal{X}_i,\mathcal{Y}_i,\Theta_i,\Pi_i)$ in Eq.~(\ref{Eq: agent representation}) and is continuously optimized using online data. Its input and output spaces, model parameters, learning mechanism, and local state transition therefore can constitute attack surfaces.
\begin{itemize}
 \item \emph{Adversarial manipulation}: an adversary may perturb inputs in $\mathcal{X}_i$ or the observations used to produce $\hat{y}_i(t)$, causing erroneous predictions or decisions. This disturbance may further pollute the difference test and extracted data and cost pairs $(\mathcal{D}_i^{*}(t), J_i(t))$ in Eq.~(\ref{eq:eq8}), thus turning the inferred time attack into harmful online updates.
 \item \emph{Model tampering/poisoning}: malicious modification of the model parameters $\Theta_i$, the learning mechanism $\Pi_i$, or the extracted dataset $\mathcal{D}_i^{*}(t)$ can corrupt the local transition $\mathcal{H}_i(\cdot)$ in Eqs.~(\ref{eq:local_update_abstract}) and (\ref{eq:state_update}). Since $\Theta_i$ is updated through the equation (\ref{eq:eq11}), the gradient of poisoning or the local target of manipulation may gradually reduce the professional ability of the agent.
 \item \emph{Evasion attacks}: the opponent may carefully design the input that leads to the wrong output, while keeping $\|\hat{y}_i(t)-y_i(t,G(t))\|$ is lower than the threshold value $\tau$ in the formula~(\ref{eq:eq8}). Therefore, the attack can circumvent novelty detection and prevent the target adaptive mechanism from being activated.
\end{itemize}

At the Interaction Layer, the principal attack surface is the localized interaction flow $\mathcal{Z}_i^{(l)}(t)$ in Eq.~(\ref{eq:interaction_flow}), which aggregates horizontal peer states and vertical functional abstractions over the dynamic topology $G(t)$.
\begin{itemize}
 \item \emph{Malicious knowledge injection}: compromised agents may transmit falsified states $s_k^{(l)}(t)$ or lower-level representations $s_j^{(l-1)}(t)$. Through the weighted aggregation coefficients $\alpha_{ik}^{(l)}$ and $\beta_{ij}^{(l-1\rightarrow l)}$ in Eq.~(\ref{eq:interaction_flow}), these signals can contaminate $\mathcal{Z}_i^{(l)}(t)$ and subsequently the extracted data and objective in Eq.~(\ref{eq:eq8}).
 \item \emph{Consensus subversion}: coordinated malicious agents may manipulate the interaction graph $G(t)$, dominate an agent's neighborhood, or bias the interaction weights $\alpha_{ik}^{(l)}$. This can distort the peer-aggregation term in Eq.~(\ref{eq:interaction_flow}), allowing adversarial information to appear as a locally consistent consensus signal $y_i(t,G(t))$.
 \item \emph{Trust exploitation}: an adversary may compromise neighbor admission or validation mechanisms and enter the neighborhood or child-agent set $\Omega_i^{(l)}$ in Eq.~(\ref{eq:containment}). Once admitted, an untrusted agent can influence both the hierarchical abstraction in Eq.~(\ref{eq:containment}) and the horizontal or vertical information flows in Eq.~(\ref{eq:interaction_flow}).
\end{itemize}

At the Regulation Layer, attacks target the global governance variables $u(t)$ and $u_i^{(l)}(t)$ in Eqs.~(\ref{eq:global_update}), (\ref{eq:local_update_abstract}), and (\ref{eq:state_update}), as well as the stability-oriented regulation mechanism in Eqs.~(\ref{eq:eq9})--(\ref{eq:eq11}).
\begin{itemize}
 \item \emph{Control-loop disruption}: an adversary may falsify the estimated local energy variation $\Delta V_i(t)$ or manipulate the regulation factor $\Gamma_i(t)$ in Eq.~(\ref{eq:eq10}). The resulting corrupted control input can alter the effective update step in Eq.~(\ref{eq:eq11}), weakening or reversing the intended regulation of local adaptation.
\item \emph{Risk-propagation exploitation}: failures originating from a compromised local state $s_i(t)$ or interaction flow $\mathcal{Z}_i(t)$ may propagate through the hierarchical composition in Eqs.~(\ref{eq:containment})--(\ref{eq:interaction_flow}). Because the global energy in Eq.~(\ref{eq:eq9}) is only locally approximated by the collection of $V_i$, coordinated local failures may also exploit cross-agent dependencies that are not captured by local monitoring.
 \item \emph{Constraint manipulation}: an adversary may corrupt the environmental feedback $\mathcal{E}(t)$, global constraints, or physical priors entering the transition models in Eqs.~(\ref{eq:global_update}), (\ref{eq:local_update_abstract}), and (\ref{eq:state_update}). Such manipulation can also bias the data-extraction function $\Psi(\cdot)$ in Eq.~(\ref{eq:eq8}), causing otherwise legitimate regulation and adaptation mechanisms to optimize toward an incorrect system objective.
\end{itemize}

These gaps stimulate a cross-layer security framework, which is structurally consistent with the three-layer architecture of Networked AI, adapts to online and unstable evolution, and can coordinate threat information across architecture boundaries. At the Agent Layer, threat modeling consistent with the architecture should solve the problems of confrontational input, indirect prompt injection, unauthorized operation, model tampering and memory poisoning~\cite{zambare2025securingagenticaithreat,deng2025aiagentsunderthreat,dong2025memoryinjection}. Separating trusted control flows from untrusted data flows can further prevent external content from changing the agent's execution logic or protected information~\cite{debenedetti2025defeating}. At the Interaction Layer, zero-trust identity governance should authenticate participating agents and enforce task-scoped communications~\cite{huang2025novelzerotrustidentityframework}. Byzantine-robust aggregation and protocol-level protection should further limit the influence of compromised agents, falsified messages, and over-privileged communication on collective learning~\cite{10657953,louck2025a2a,louck2025protocolsecurity}. At the Regulation Layer, trusted monitoring, anomaly isolation, fallback, and rollback should prevent falsified local signals and agent-level failures from propagating into system-wide instability. Because static defenses may fail against adaptive adversaries, the resulting framework should be evaluated under adaptive attacks rather than fixed attack patterns~\cite{zhan2025adaptive}. Together, these mechanisms of cross-layer and resilient security are being developed to jointly form a unified defense across the life-cycle of Networked AI, contributing to a trusted online evolutive system.

\section{Paradigm Migrations}

The development of Networked AI is closely related to many research directions such as large-scale foundation models, distributed intelligence, continual learning, multi-agent collaboration and emergency collective systems. Although these fields are largely independent, they together provide an important technical basis for Networked AI to operate in an open and dynamic environment (Table~\ref{tab:evolution}).
\subsection{Foundation Models and Standalone AI Systems}

The latest progress of deep learning is mainly driven by independent large-scale training models. AlexNet~\cite{krizhevsky2012imagenet}, ResNet~\cite{he2016deep}, Transformer~\cite{NIPS2017_3f5ee243} lay the foundation for modern AI systems.
Subsequent developments in large-scale language models and multimodal learning, including GPT series models~\cite{GPT_3}, CLIP~\cite{CLIP} and PaLM~\cite{chowdhery2023palm}, show excellent ability in language understanding, reasoning and generative intelligence.

Despite the success of standalone AI systems, they are increasingly limited in terms of scalability, adaptability and deployment efficiency, and the improvement of performance relies heavily on the exponential increase in computing resources and training data. In addition, offline pre-training paradigms often struggle in non-stationary environments that require continual adaptation, application-specific customization and online interaction.

These limitations promote the transition from standalone models to distributed and evolving intelligent systems.

\subsection{Distributed AI and Multi-Agent Systems}

Distributed artificial intelligence (DAI) and Multi-Agent Systems (MAS) provide an early theoretical basis for distributed intelligent systems~\cite{wooldridge2009introduction,dorri2018multiagent}. In these examples, multiple autonomous agents collaborate through coordination mechanisms to solve complex tasks that are difficult for a single agent to complete.

The existing research mainly focuses on decentralized optimization, consensus dynamics, task allocation and game theory coordination~\cite{cao2013overview,zhang2018fully}. In recent years, distributed multi-agent reinforcement learning (MARL) systems~\cite{busoniu2008comprehensive,zhang2021multi} has received wide attention in the large-scale autonomous system. These methods provide important insights for distributed decision-making and local-to-wide coordination.

However, most existing MAS and MARL frameworks mainly focus on predefined collaboration goals in a relatively static environment. Problems such as online evolutive adaptation, autonomous ability growth, dynamic structure reconfiguration and large-scale heterogeneous intelligence have not been fully studied.

\subsection{Federated and Distributed Learning}

Federated Learning (FL) introduces a distributed learning paradigm in which multiple Agents collaborate to train models without directly sharing raw data~\cite{mcmahan2017communication,li2020federated,liu2025privacy,ribero2024federated}.
FL has significantly improved the ability of privacy protection and distributed deployment, making it highly relevant to edge intelligence and large-scale networked systems.

Recent research has extended FL to distributed and asynchronous settings~\cite{lalitha2019peer,sun2021decentralized}, personalized federated learning~\cite{tan2022personalized,xu2024cooperative}, and federated reinforcement learning~\cite{zhu2023federated,wagle2025multi}.
These studies involve communication efficiency, non-IID data distribution and distributed optimization stability.

However, federated learning mainly focuses on collaborative parameter aggregation, rather than autonomous system evolution. Most FL systems still rely on relatively fixed model structures and predetermined training goals, and lack mechanisms for continual adaptation, structural self-reorganization and sudden collective intelligence.

\subsection{Continual Learning and Online Adaptation}

Continual learning aims to enable AI systems to constantly acquire new knowledge while maintaining their previous learning ability~\cite{parisi2019continual,mundt2023wholistic}.
The research on lifelong learning, online adaptation and continual reinforcement learning investigated catastrophic forgetting, knowledge transfer and long-term adaptation~\cite{chen2018lifelong,khetarpal2022towards,liu2025crcl}.

Recent studies have begun to explore distributed continual learning and collective lifelong intelligence~\cite{soltoggio2024collective}. In particular, Shared-Experience Lifelong Learning (ShELL) system proposes collaborative knowledge accumulation between distributed agents operating at the edge. These advances show that distributed intelligent systems can achieve continual adaptation through experience sharing.

However, existing continual learning frameworks generally focus on isolated learning processes or small-scale collaborative systems, with the target problem primarily on adaptation for generalization, instead of the evolution for applications as in Networked AI. Key missing parts can include a unified framework integrating distributed coordination, autonomous data generation, online structural evolution, and system-level regulation.

\subsection{Collective Intelligence and Emergent Systems}

Collective intelligence and swarm intelligence study how global intelligent behaviors emerge from local interactions among distributed agents~\cite{bonabeau1999swarm,kennedy2001swarm}. 
Research in complex adaptive systems, self-organization, and emergent communication has demonstrated that decentralized interactions can generate sophisticated system-level behaviors~\cite{cao2018emergent}.

These ideas have inspired important developments in swarm robotics, distributed optimization, and cooperative AI~\cite{conitzer2023foundations}. 
More recently, emergent collaborative reasoning among learning agents have become active research topics in large-scale AI systems~\cite{cao2018emergent}.

Nevertheless, the foundations of emergent intelligence in large-scale AI networks remain insufficiently understood. Key missing problems can include the formal characterization of collective intelligence, stability of evolving interaction structures, and the relationship between local adaptation and global system behavior.

\begin{table*}[t]
\centering
\caption{Paradigm Migrations Toward Networked AI.}
\label{tab:evolution}
\renewcommand{\arraystretch}{1.3}
\begin{tabular}{lccccc}
\hline
\textbf{Paradigm}
&
\textbf{Intelligence Source}
&
\textbf{Data Source}
&
\textbf{Adaptation}
&
\textbf{Coordination}
&
\textbf{Scalability}
\\
\hline

Expert Systems
&
Human Rules
&
Knowledge Bases
&
None
&
None
&
Low
\\

Deep Learning
&
Model Parameters
&
Labeled Datasets
&
Offline
&
None
&
Moderate
\\

Foundation Models
&
Large-scale Pretraining
&
Web-scale Data
&
Limited
&
None
&
Compute-Bounded
\\

Multi-Agent AI
&
Agent Cooperation
&
Environment Interaction
&
Partial
&
Explicit
&
Moderate
\\

Networked AI
&
Collective Evolution
&
Knowledge and   Operation Data&
Online Evolutive&
Adaptive + Regulated
&
Hierarchical and Near-linear\\

\hline
\end{tabular}
\end{table*}

\section{Application Frontiers}

Now, we will study how the Networked AI architecture is mapped to several application areas where data and intelligent efficiency have become major bottlenecks, and discuss what layered design provides in each situation.

\subsection{Intelligent Internet of Things}

The deployment of the Internet of Things across ecological surveillance, smart infrastructure, and industrial sensing produces heterogeneous, noisy and rarely static data flows. In practice, what complicates these data streams is not so much the perceived uncertainty itself as the operational constraints surrounding it: tight energy budgets, intermittent connections, and local data governance rules that limit the possible dissemination of raw data. It is difficult for independent examples to adapt to such a setting. Streaming the original measurement data to the cloud is expensive in terms of bandwidth and energy, and retraining each site in isolation will blind the system to cross-site drift and events often observed by few single nodes. Networked AI regards each edge node as an autonomous Agent-Layer unit, which perceives, filters and adapts locally, and only shares bounded summaries with its neighbors.

On-site sonar monitoring of wild river ecosystems is a typical example. Under the sparse network coverage, the moving targets of the underwater target must be tracked~\cite{SALINA}. Agents who handle these streams can be responsible for fish counting, trajectory tracking or abnormal detection, and they do not require a global view. The governance dimension is equally binding: community-led ecological projects, infrastructure operators and industrial facilities are often unwilling or legally unable to gather raw data, parameters or embedding. By exchanging compact prototypes and sound retrieval signals, agents can compare observations across domains, while the underlying data remains local~\cite{FUSED}.

The signal processing tool specifically instantiates the Interaction Layer. Kalman filtering, temporal smoothing, background modeling and graph signal processing exceeding $G(t)$ allow each agent to suppress noise before sharing any content and only retain evidence related to the task. On top of this, the Regulation Layer adjusts the update steps through $\Gamma_i(t)$, and reallocates sensing, computation and energy when the network degrades or the environment changes suddenly. This brings specific rather than generic: edge analysis within the scope of energy and connectivity budgets, and the exchange protocol for maintaining local management data.




\subsection{Embodied Intelligence}

Embodied systems such as autonomous robots, drones and collaborative manipulators run in a physically dynamic and partially observable environment, in which the strategy of offline training will be downgraded once it deviates from the training distribution. Centralized control makes the problems in multi-agent settings more complicated, because communication delays and environmental changes leave little room for anything other than device adaptation. The Interaction Layer directly solves this problem: once the difference $\|\hat{y}_i(t) - y_i(t, G(t))\| > \tau$ appears in the physical interaction process, it folds the a priori in the fields of kinematic limit and contact dynamics as the cost function $J_i $. Data efficiency is decisive here, because under real physical conditions, high-quality interactive data is scarce and expensive.

The multimodal case expands this picture. The agent for perception through visual, audio, touch and proprioception requires the Agent Layer to host vision-language-action models and lightweight edge perception, all of which represent $(\mathcal{X}_i, \mathcal{Y}_i, \Theta_i, \Pi_i)$. Then, cross-modal alignment becomes the responsibility of the Interaction Layer: adaptive filtering coordinates the asynchronous stream into a consistent representation, so that language instructions can be transformed into action sequences based on the actual content of the agent's perception. Human-robot interaction is the most difficult to emphasize this cycle, because scenario understanding, voice commands and tactile feedback must be coordinated under a tight delay budget.

The collaboration between agents is the most direct contribution of the network structure. Task-related experience flows along the dynamic diagram $G(t)$, so skill acquisition accelerates without any synchronous global updates, and the Regulation Layer processes the formation again when the agent fails or the task is reassigned through the Lyapunov-based actor $\Gamma_i(t)$. Configuration and load balancing. The modular structure brings a practical consequence: in security-critical deployments, failures can be attributed to specific agents and their audited behavior, and a single strategy cannot provide such transparency.

\subsection{Spatial Intelligence}

Spatial distribution perception, including synchronous positioning and mapping, collaborative scene reconstruction and multi-viewpoint object recognition, puts forward consistency and bandwidth requirements that are difficult to meet by standalone methods as the scene grows. Each agent only observes a part of the environment, so a coherent global map must be combined from these local views. Networked AI organizes the participating perception subjects on an adaptive interaction diagram, and verifies and integrates its observation results through the knowledge exchange protocol of the Interaction Layer. Consistency is a concern of the Regulation Layer: it enforces geometric and semantic constraints between agents so that contradictory local mappings do not damage the shared scenario pictures. Due to the different abilities of agents, in the case of a LiDAR-based depth estimation and another visual-language semantic labeling, they can form task-specific subnetworks according to the hierarchy of Eq.~(\ref{eq:containment}), and the generated topology and Trace coverage and bandwidth, not pre-fixed.

The second requirement is about the temporal consideration. The real environment will not stand still: the map is outdated, the observation value conflicts, and the agent population itself moves, so the system must be constantly updated, instead of converging and staying unchanged. Therefore, the joint evolution of agent representation and interaction structure becomes a defining attribute of large-scale spatial intelligence, not an accidental feature, which is most important when coverage, uncertainty and resources change during operation. Compared with the overall mapping that relies on global aggregation, the advantage here is to maintain consistency under changes while reducing data and calculations.

\subsection{Industrial Intelligence}

Industrial AI includes a tightly coupled process across edge devices, factory controllers, and cloud infrastructure, in which delay-sensitive control decisions must coexist with long-horizon optimization goals. Conventional standalone AI schedulers lack the flexibility to respond to equipment failures, demand fluctuations or supply chain interruptions without the need for expensive retraining cycles. Networked AI realizes edge-cloud collaborative learning by distributing decision-making agents at the industrial hierarchy. Each agent interactively adapts to application-specific operational feedback, and the Regulation Layer maintains global life through the regulated update rules of Eq.~\eqref{eq:state_update} Target production and implement security constraints.

Dynamic scheduling optimization exemplifies the advantages of the networked AI paradigm: role-based agent communities representing individual production cells or operators who negotiate resource allocation or operational flows through the Interaction Layer, guided by application-specific priors encoding process physics and quality specifications. This self-organizing behavior reduces dependence on centralized planners, and enables graceful degradation under partial system failures, mimicking exactly how human organization operates in vertical industries. As no existing industry can offer the massive high-quality data as required by large standalone models, Networked AI offers the industries an application-specific, intelligent and autonomous system that can be not only developed upon scarce and private industrial data, but also continuously optimized for target scenarios during its operation.

Moreover, by transforming industrial AI from a black-box optimizer into an interpretable distributed decision network, Networked AI improves both operational efficiency and system accountability, providing operators with traceable, layer-wise explanations of scheduling decisions and anomaly responses consistent with the gray-box architecture, as established in Section III-E.

%
\subsection{Cellular Networks and Beyond}  

Cellular networks are one of the most specific industrial settings for Networked AI, because they are already composed of a large number of distributed entities, which must coordinate under changing radio, traffic and service conditions. Such networks are built on heterogeneous infrastructure with emerging and legacy components, while serving non-fixed channels, uneven and sudden needs, mobile users, and strict latency and reliability goals. In this case, it is difficult to keep the information obtained through offline training or centralized adjustment up to date after deployment. Base stations, user equipment (UE), edge computing platforms, core-network analytics functions such as the Network Data Analytics Function (NWDAF), and operations, administration, and maintenance (OAM) systems can each be regarded as an operational agent that observes a local state and acts under network-wide service and operational policies. These entities form the Agent Layer, and recent 5G-Advanced studies on AI/ML for the NR air interface and on NG-RAN intelligence provide an initial standardization basis for this evolution~\cite{3gpp2024ai,lin2023embracing}.

Cellular AI spans radio-level optimization and network-level automation. Learning-based methods have been studied for channel state information feedback, channel prediction, beam management, positioning, and related air-interface functions~\cite{3gpp2024ai,guo2022overview}. AI can also support traffic forecasting, energy-saving control, mobility and load balancing, fault diagnosis, and slice or service assurance~\cite{3gpp2022data,coronado2022zero}. The role of the Interaction Layer organizes telemetry, radio measurements, service indicators, and model information across the radio access network, core, edge, and terminals. When the difference between the predicted network behavior and the measured observed value exceeds the threshold, for example, $||\hat{y}_i(t) - y_i(t, G(t))|| > \tau$, selective local adaptation can be triggered, and only the relevant agent or service area can be updated.

 
Cellular infrastructure can support edge inference, federated learning, model distribution, and cloud–edge–device coordination~\cite{letaief2022edge,ji2024computational}. This is meaningful only when AI services are delivered with predictable performance, privacy protection, and manageable cost. The Regulation Layer addresses this requirement through service-level assurance, slice policy enforcement, safety validation, anomaly isolation, fallback, and rollback before AI decisions or model updates are widely applied. In this process, regulation operators such as $\Gamma_i(t)$ can constrain local updates under global stability and service requirements. 

In addition to terrestrial cellular networks, Orbital AI is emerging, and AI workloads are directly deployed in the sky-based computing infrastructure, especially the Low Earth Orbit (LEO) satellite constellation. The satellite can perform local inference, intermediate feature extraction, task unloading and model coordination across satellite links, instead of continuously downlinking the original earth observation, environment or communication data to the ground cloud. This will convert the LEO giant constellation from communication relay to a distributed orbital computing platform in the 6G era~\cite{OrbitTransit,10409745}. The future Orbital AI system can further integrate the federal world arrangement. In such a framework, LEO, MEO (Medium Earth Orbit, commonly 5,000–20,000 km above Earth), GEO (Geostationary Earth Orbit at about 35,786 km above the equator) and ground cloud resources cooperate as a unified computing continuum. Recent work on AI-empowered non-terrestrial networks, in-orbit machine learning payloads, and satellite edge intelligence suggests that the architecture of Networked AI will become central to future 6G infrastructure~\cite{10409745,In_Orbit,10887238}. Orbital AI in space is therefore not merely AI deployed on satellites; it is an adaptive, radiation-aware, and energy-conscious computing paradigm designed for autonomous Networked AI beyond Earth.

\section{ Industrial Impacts and Open Challenges}

\subsection{Impact on 6G and Applications} 
Networked AI offers an organizing framework for 6G systems, in which the network is expected to evolve from a connectivity infrastructure into one that also coordinates distributed AI functions across devices, base stations, edge clouds, core-network functions, and application platforms. The IMT-2030 framework identifies “\textit{AI and Communication}” as one of its usage scenarios and treats ubiquitous intelligence as an overarching aspect of future systems~\cite{itur2023framework}. As a preceding effort, the NSAI research community also explicitly identified the roadmap of "\textit{AI as a Networking System, and Network as an AI system}", which virtually equals network and AI~\cite{nsaisong}. These directions map naturally onto the three-layer Networked AI architecture for potentially universal 6G applications: heterogeneous models, edge and terminals, network functions, and intelligent application/service functions constitute the Agent Layer; cross-domain data, context, and model information are exchanged and optimized through the Interaction Layer; and validation, policy enforcement, fallback, rollback, and accountability are handled by the Regulation Layer. For 6G and its applications, the question is not whether AI can be introduced, but how AI functions can be deployed, updated, and governed at network scale.

Networked AI has also transformed several research topics into specific design problems of industrial standardization and deployment. The integration of perception, communication, computing and AI requires agent to combine multi-modal observation with distributed decision-making, while maintaining a controlled state of data and model life cycles. Energy efficiency and sustainability may benefit from online evolutive learning, because in principle, model updates can be limited to units, equipment or service areas where traffic volume, radio conditions or tasks actually change, thus avoiding network-wide retraining. When local learning, cross-domain interaction and system-level regulation are clearly separated, credibility and security problems can be solved more systematically. Network digital twins can also be used as a controlled environment to verify and stress-testing AI policies before they are applied to live networks~\cite{tao2024wireless}.

In summary, from the perspectives of Information and Communication Technology (ICT) industry, we observe a ramping-up momentum that \textit{6G is being built by AI and for AI.} Particular for 6G applications, the value of Networked AI lies less in adding AI algorithms to individual devices than in providing a managed infrastructure for intelligent services. Typical vertical applications, including intelligent IoT, embodied, spatial, and industrial intelligence, require more than connectivity. They depend on coordinated intelligence among devices, edge nodes, network functions, and application platforms, supported by predictable latency, trusted data exchange, and accountable decision-making~\cite{letaief2022edge}. The impact converges with the aforementioned NSAI paradigm: AI is expected to become a native capability for operating and optimizing 6G networks, while 6G provides a deployable and governed infrastructure for AI services at industrial scale. And therefore, \textit{6G applications are virtually Networked-AI applications, }which in-turn are expected to drive the evolution of the entire ICT industry to its next generation. 

\subsection{Key Open Challenges}


Despite its potential to enable scalable and adaptive distributed intelligence, Networked AI introduces several fundamental open challenges spanning the agent, interaction, and regulation layers. These challenges arise from the need to jointly support heterogeneous intelligence, decentralized evolution, and system-level control in dynamic environments.

At the Agent Layer, one of the main challenges is to develop heterogeneous intelligent agents that can continuously adapt under resource constraints. Unlike the traditional distributed learning paradigm, Networked AI must support agents with different architectures, modalities and learning paradigms. This raises important theoretical questions about the compatibility, the transferability of learned knowledge, and the convergence under heterogeneous local objectives. The difficulties further introduced in a nonstationary environment may include catastrophic forgetting. From an application perspective, edge devices, autonomous robots, wearable systems and smart sensing platforms require lightweight yet reliable intelligence to run under limited computation, energy and communication budgets. Designing and ensuring the robustness, interpretability, privacy preservation and trustworthy model of agentic AI community is still an open research direction.

The Interaction Layer raises other challenges related to decentralized coordination and collective evolution. A central problem is to design adaptive interactive topologies that can evolve according to task requirements, environmental changes and resource availability. The existing distributed optimization and consistency framework is usually not suitable for highly dynamic and heterogeneous systems characterized by asynchronous updates, communication delays and partial participation. Another challenge involves balancing local specialization and global consistency. Excessive synchronization will inhibit diversity and adaptability, while insufficient coordination will lead to instability and fragmentation of learning behavior. In addition, understanding the emerging collective intelligence has not been explored to a large extent, especially in systems where interaction structures and proxy behaviors have evolved over time. Robustness against adversarial agents, faulty communications, and cascading error propagation is also critical for large-scale deployments.

In the Regulation Layer, the main challenge is to achieve system-level stability, controllability and reliability while maintaining decentralized autonomy. Networked AI requires a hierarchical control mechanism, which can regulate a large number of adaptive agents through global goals, constraints and feedback signals. Providing strict theoretical guarantees for the convergence, stability and safety of continuously evolving distributed systems is still an open issue. In addition, dynamic resource arrangement across computing, communication and energy dimensions is crucial for scalable operations. Regulatory mechanisms must also address ethical and societal concerns, including fairness, accountability, transparency and governance of autonomous collective behavior. These problems have become particularly important in critical applications such as health care, smart grids, autonomous transportation, and national or industrial-scale cyber-physical systems.

Beyond the individual layers, several cross-layer challenges remain unresolved. A unified theoretical framework capable of jointly modeling local intelligence, interaction dynamics, and global regulation is still at its early stage. Scalability to massive networks with millions of adaptive agents, resilience against failures and attacks, and interoperability across heterogeneous platforms remains to be proven by researchers in both academia and industry. Ultimately, the central challenge of Networked AI is to reconcile four competing objectives simultaneously: local autonomy, collective intelligence, continual evolution, and globally regulated stability. Addressing these challenges will refine the future theoretical foundations and real-world applicability of Networked AI.

\section{Conclusion}

Networked AI offers the potential for substantially improved data and intelligence efficiency through localized interaction, hierarchical factorization, and system-level regulation, and the foreseeable and explainable AI that can be working reliably with mankind. In this paper, we have established a three-layer hierarchical architecture of networked AI, together with a unified mathematical formalism encompassing agent state dynamics, interaction-driven signal processing, and global stability control, where the three fundamental transitions are highlighted. 

Training large models with small data: near-linear vs. exponential growth of resource demand. Conventional standalone AI systems depend on massive offline datasets whose required volume can grow exponentially with system capability or task complexity. Networked AI replaces this with continuously generated, interaction-driven high-quality data in near-linear scaling. This transition shifts AI systems from data-hungry monolithic or offline training towards sustainable collective evolution, representing a fundamental improvement in \textit{data efficiency}. 

From generalization to application-specific adaptation: online evolution vs. offline training. Traditional models are trained once and deployed statically, struggling under distributional shift and out-of-distribution novelties. Networked AI's Interaction Layer enables online detection of statistical discrepancies through adaptive signal processing, triggering targeted, localized application customization, i.e., online evolution, without global retraining. Smaller and agile models are therefore possible, representing a fundamental improvement in \textit{intelligence efficiency} . 

Achieving interpretability through networking: gray-box vs. black-box. Deep learning's black-box nature limits controllability and trust, whereas networked AI transforms this by structuring intelligence as explicitly defined layers with explainable interfaces: signal processing kernels ground the Interaction Layer in provable estimation theory, and Lyapunov-based control provides formal stability conditions for the Regulation Layer. This layered decomposition renders the system an \textit{interpretable and amenable} gray-box without sacrificing the representational power of deep models. 

New quantitative metrics of data efficiency, intelligence efficiency, and interpretability remain to be discussed and defined in the research and technical communities. Nevertheless, these three transitions together are defining Networked AI not only as an engineering approach with prolific prospects, but also as a paradigm shift, transforming from isolated, data-saturated models toward collectively intelligent, resource-efficient, and interpretable systems, autonomously evolving for the target application scenarios. This transformative paradigm is being seamlessly bridged with the rapidly evolving landscape of large-scale agentic and self-improving AI systems.

\appendices
\section{Scaling Laws and Remarks}
\label{app:scaling_theorems}

Let $N$ be the number of heterogeneous task components, $c>1$ the feature-expansion factor, and $R(N)$ and $C(N)$ the sample and computational requirements for reaching a prescribed performance criterion.

\begin{theorem}[Standalone scaling under global coupling]
\label{thm:standalone_scaling}
Consider a standalone end-to-end system that jointly models $N$ heterogeneous task components. Suppose that: (i) each component contributes at least $c$ distinguishable feature configurations; (ii) no structural factorization excludes cross-component combinations; and (iii) resolving each relevant joint configuration requires at least a constant number of samples and operations. Then
\begin{equation}
R_{\mathrm{s}}(N)=\Omega\!\left(c^N\right),
\qquad
C_{\mathrm{s}}(N)=\Omega\!\left(c^N\right).
\label{eq:standalone_bound}
\end{equation}
\end{theorem}

\begin{proof}
Let $\mathcal{X}_i$ be the feature-configuration set of component $i$, where $|\mathcal{X}_i|\geq c$. Without structural factorization, the
joint configuration space is
\begin{equation}
\mathcal{X}_{\mathrm{s}}
=\prod_{i=1}^{N}\mathcal{X}_i,
\qquad
|\mathcal{X}_{\mathrm{s}}|
=\prod_{i=1}^{N}|\mathcal{X}_i|
\geq c^N.
\label{eq:joint_space}
\end{equation}
If every relevant configuration requires at least $q_{\mathrm{s}}>0$ samples and $r_{\mathrm{s}}>0$ operations per sample, then
\begin{equation}
R_{\mathrm{s}}(N)
\geq q_{\mathrm{s}}c^N,
\qquad
C_{\mathrm{s}}(N)
\geq r_{\mathrm{s}}R_{\mathrm{s}}(N).
\label{eq:standalone_resource}
\end{equation}
Therefore,
$R_{\mathrm{s}}(N),C_{\mathrm{s}}(N)=\Omega(c^N)$.
\end{proof}

\begin{theorem}[Networked-AI scaling under hierarchical factorization]
\label{thm:networked_scaling}
Consider $N$ agents organized into a balanced hierarchy with branching factor $b>1$ and depth
\begin{equation}
L=\left\lceil\log_b N\right\rceil.
\label{eq:hierarchy_depth}
\end{equation}
Suppose that: (i) every hierarchical level increases the feature-combination complexity along an agent-centered path by at most a factor $c$; (ii) interaction and optimization remain localized, avoiding an unrestricted Cartesian product across different paths; and (iii) the samples and operations per locally processed feature combination are bounded by constants independent of $N$. Then
\begin{equation}
\begin{aligned}
R_{\mathrm{n}}(N),C_{\mathrm{n}}(N)
&=O\!\left(Nc^{\lceil\log_bN\rceil}\right)\\
&=O\!\left(N^{1+\log_b c}\right).
\end{aligned}
\label{eq:networked_bound}
\end{equation}
\end{theorem}

\begin{proof}
Along any agent-centered path, at most $c^L$ feature combinations are generated after $L$ hierarchical levels. With at most $N$ such paths
and no unrestricted cross-path expansion, the aggregate number of locally processed feature combinations satisfies
\begin{equation}
B_{\mathrm{n}}(N)
\leq K_0Nc^L
\label{eq:feature_burden}
\end{equation}
for some constant $K_0>0$ independent of $N$. If each locally processed combination requires at most $q_{\mathrm{n}}>0$ samples and
$r_{\mathrm{n}}>0$ operations, then
\begin{equation}
R_{\mathrm{n}}(N)
\leq q_{\mathrm{n}}B_{\mathrm{n}}(N),
\qquad
C_{\mathrm{n}}(N)
\leq r_{\mathrm{n}}B_{\mathrm{n}}(N).
\label{eq:networked_resource}
\end{equation}
Hence,
\begin{equation}
R_{\mathrm{n}}(N),C_{\mathrm{n}}(N)
=O\!\left(Nc^L\right).
\label{eq:networked_intermediate}
\end{equation}
Moreover,
\begin{equation}
c^{\lceil\log_bN\rceil}
\leq c^{1+\log_bN}
=cN^{\log_b c}.
\label{eq:ceiling_bound}
\end{equation}
Because $c$ is independent of $N$, substituting
$L=\lceil\log_bN\rceil$ into Eq.~\eqref{eq:networked_intermediate}
yields Eq.~\eqref{eq:networked_bound}.
\end{proof}

\begin{corollary}[Scaling separation]
\label{cor:scaling_separation}
Under the assumptions of
Theorems~\ref{thm:standalone_scaling} and~\ref{thm:networked_scaling},
\begin{equation}
\frac{R_{\mathrm{s}}(N)}{R_{\mathrm{n}}(N)},
\frac{C_{\mathrm{s}}(N)}{C_{\mathrm{n}}(N)}
=\Omega\!\left(
\frac{c^N}{Nc^{\lceil\log_bN\rceil}}
\right).
\label{eq:scaling_ratio}
\end{equation}
Thus, hierarchical factorization changes the resource dependence on $N$ from exponential under unrestricted global coupling to $O\!\left(N^{1+\log_b c}\right)$. When $\log_b c\ll1$, the exponent approaches one; we refer to this regime as \emph{near-linear scaling}.
\end{corollary}

\begin{remark}[Scope]
Theorem~\ref{thm:standalone_scaling} is conditional on unrestricted global coupling and is not a universal lower bound for all monolithic
architectures. Theorem~\ref{thm:networked_scaling} is an upper bound under localized hierarchical factorization. In particular, $c<b$ gives a subquadratic bound, whereas near-linear scaling requires the stronger regime $\log_b c\ll1$.
\end{remark}

\begin{remark}[Connection to the experiments]
Figure~\ref{fig:task_agent_number_boxplot} evaluates $N\in\{2,3,4,5,6\}$. Over this finite range, $N^{1+\log_b c}$ can become $c\cdot N$ when $b$ is configured to $N$. The observed linear fit is therefore consistent with Theorem~\ref{thm:networked_scaling}. Figures~\ref{fig:model_scale_boxplot} and~\ref{fig:task_complexity_boxplot} provide complementary evidence of reduced empirical resource sensitivity with respect to model and application scales, as related to $c$; they do not directly show the bounds with respect to $N$.
\end{remark}

\section{Conditional Stability of the Regulation}
\label{app:stability}

Let $g_i(t)=\nabla_{\Theta_i}J_i(\mathcal{D}_i^{*}(t),\Theta_i(t))$ and
$a_i(t)=\eta\Gamma_i(t)$ for notation simplicity here, so that Eq.~(\ref{eq:eq11}) can be written as
$\Theta_i(t+1)=\Theta_i(t)-a_i(t)g_i(t)$. We represent the approximation
in Eq.~(\ref{eq:eq9}) as
\begin{equation}
V(S(t))=\sum_{i=1}^{K}V_i(\Theta_i(t),\mathcal{Z}_i(t))+\varepsilon(t),
\end{equation}
where $\varepsilon(t)$ collects cross-agent coupling and decomposition
residuals. For the stability analysis, the non-parameter components of $s_i(t)$ are absorbed into the residual $\varepsilon(t)$.

\textbf{Assumption 1:}
The global energy function $V$ is lower bounded, and each $V_i$ has an $L_i$-Lipschitz gradient. Moreover,
for some $\mu_i>0$, $\overline{\Delta}_i<\infty$, and $\rho_i\geq0$,
\begin{align}
\langle\nabla_{\Theta_i}V_i(t),g_i(t)\rangle
&\geq\mu_i\lVert g_i(t)\rVert^2, \label{eq:inequity1}\\
\Delta V_i(t)&\leq\overline{\Delta}_i, \label{eq:inequity2}\\
\varepsilon(t+1)-\varepsilon(t)
&\leq\sum_{i=1}^{K}\rho_i a_i(t)\lVert g_i(t)\rVert^2,\label{eq:inequity3}
\end{align}
where $0<\eta<\min_i 2\mu_i/L_i$ and
$\rho_i<\mu_i-L_i\eta/2$.

These assumed conditions are practically plausible in controlled Networked-AI deployments rather than being automatic properties of arbitrary networks. Local smoothness is commonly satisfied, or can be enforced locally, when agent objectives use smooth losses and bounded parameter updates. The first inequality Eq. (\ref{eq:inequity1}), the descent-alignment condition, can be promoted by augmenting each task objective with a stability regularizer derived from $V_i$, or by projecting task gradients onto a descent-compatible direction. The second inequality  Eq. (\ref{eq:inequity2}), the bounded-variation condition is reasonable when inputs, gradients, and candidate updates are clipped or otherwise constrained, which prevents $\Gamma_i(t)$ from vanishing. The third inequality  Eq. (\ref{eq:inequity3}), the coupling-residual bound is compatible with the localized architecture when neighborhood sizes, interaction weights, communication delays, and environmental drift remain bounded. The learning-rate and residual-margin conditions afterward can then be checked or enforced online by conservative step-size selection and monitoring of the estimated descent margin. 

\begin{theorem}[Conditional Stability]\label{thm:stability}
Under Assumption 1, the regulated dynamics satisfy
\begin{equation}\label{eq:descent}
V(S(t+1))-V(S(t))
\leq-\sum_{i=1}^{K}a_i(t)c_i\lVert g_i(t)\rVert^2\leq0,
\end{equation}
where $c_i=\mu_i-L_i\eta/2-\rho_i>0$. Consequently,
$V(S(t))$ converges and $\lVert g_i(t)\rVert\rightarrow0$ for every agent.
If the resulting invariant stationary set contains a unique equilibrium
$S^{\star}$ and $V$ is positive definite with respect to $S^{\star}$, then
$S^{\star}$ is asymptotically stable.
\end{theorem}

\begin{proof}
By the descent lemma and Assumption 1,
\begin{align}
V_i(t+1)-V_i(t)
&\leq-a_i(t)\langle\nabla_{\Theta_i}V_i(t),g_i(t)\rangle
+\frac{L_i}{2}a_i^2(t)\lVert g_i(t)\rVert^2 \\
&\leq-a_i(t)\left(\mu_i-\frac{L_i\eta}{2}\right)
\lVert g_i(t)\rVert^2,
\end{align}
because $0<a_i(t)\leq\eta$. Summing over $i$ and applying the residual
bound gives the stated descent inequality Eq. (\ref{eq:descent}). Since Eq.~(\ref{eq:eq10}) and
$\Delta V_i(t)\leq\overline{\Delta}_i$ imply
$a_i(t)\geq\eta\exp(-\kappa\overline{\Delta}_i)>0$, summing the descent inequality over time yields $\lVert g_i(t)\rVert\rightarrow0$. The final stability claim follows under the additional conditions stated above.
\end{proof}

\bibliographystyle{IEEEtran}
\bibliography{references}

\end{document}